\documentclass[10pt,twocolumn,letterpaper]{article}

\usepackage[preprint]{cvpr} 
\usepackage[skins]{tcolorbox}
\definecolor{best}{RGB}{253, 104, 100}
\definecolor{second}{RGB}{255, 204, 153}

\newtcbox\best{hbox, on line, colback=best, enhanced, frame hidden, boxrule=0pt, top=-2pt, bottom=-2pt, right=-2pt, left=-2pt, sharp corners}
\newtcbox\second{hbox, on line, colback=second, enhanced, frame hidden, boxrule=0pt, top=-2pt, bottom=-2pt, right=-2pt, left=-2pt, sharp corners}

\definecolor{cvprblue}{rgb}{0.21,0.49,0.74}
\usepackage[pagebackref,breaklinks,colorlinks,allcolors=cvprblue]{hyperref}

\usepackage{amsmath}

\usepackage{bm}
\usepackage[skins]{tcolorbox}
\usepackage{multirow}
\usepackage{algorithm}
\usepackage{algpseudocode}
\usepackage{dsfont}

\title{D\textsuperscript{3}-Human: Dynamic Disentangled Digital Human from Monocular Video}

\author{
    \large Honghu Chen \quad Bo Peng \quad Yunfan Tao \quad Juyong Zhang\thanks{Corresponding author} \vspace{0.5 mm}\\
    {\normalsize University of Science and Technology of China}\\
    {\tt\small \{honghuc@mail., ustcpbb@mail., taoyf@mail., juyong@\}ustc.edu.cn} \\
}

\begin{document}

\twocolumn[{%
\renewcommand\twocolumn[1][]{#1}%
\maketitle
% \vspace{-9mm}
\includegraphics[page=1,width=\textwidth]{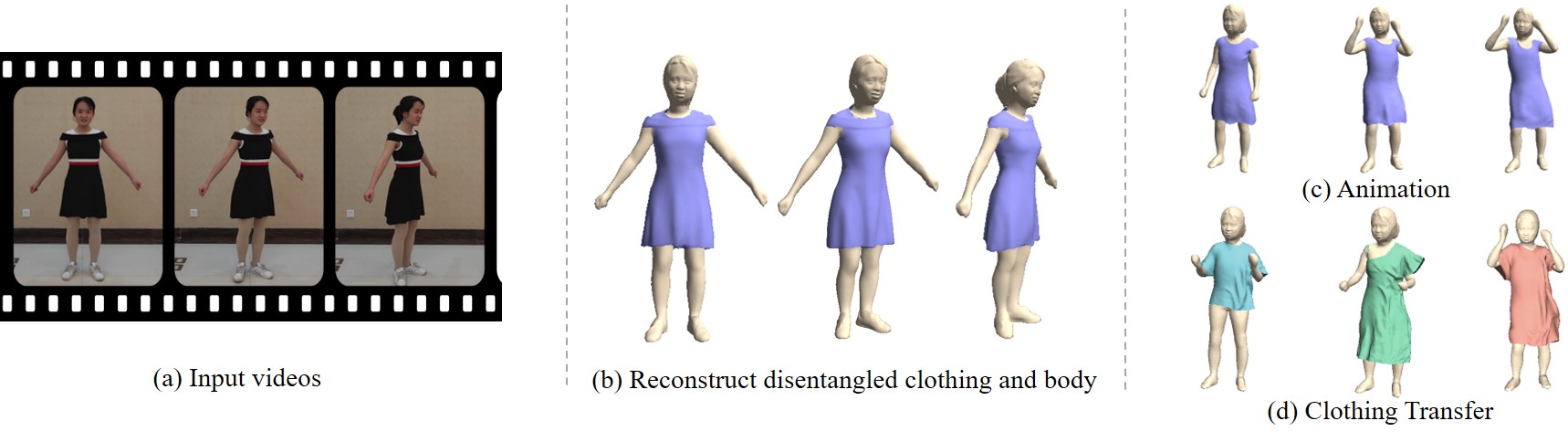}
\vspace{-3mm}
\captionof{figure}{D\textsuperscript{3}-Human can (b) reconstruct disentangled clothing and body from (a) input video, enabling (c) animation and (d) clothing transfer after reconstruction.
Project page: \href{https://ustc3dv.github.io/D3Human/}{https://ustc3dv.github.io/D3Human/}.
}
\label{fig:teaser}
\vspace{3mm}
}]

\begin{abstract}
We introduce D\textsuperscript{3}-Human, a method for reconstructing \textbf{D}ynamic \textbf{D}isentangled \textbf{D}igital Human geometry from monocular videos. 
Past monocular video human reconstruction primarily focuses on reconstructing undecoupled clothed human bodies or only reconstructing clothing, making it difficult to apply directly in applications such as animation production. 
The challenge in reconstructing decoupled clothing and body lies in the occlusion caused by clothing over the body. To this end, the details of the visible area and the plausibility of the invisible area must be ensured during the reconstruction process.
Our proposed method combines explicit and implicit representations to model the decoupled clothed human body, leveraging the robustness of explicit representations and the flexibility of implicit representations. 
Specifically, we reconstruct the visible region as SDF and propose a novel human manifold signed distance field (hmSDF) to segment the visible clothing and visible body, and then merge the visible and invisible body.
Extensive experimental results demonstrate that, compared with existing reconstruction schemes, D\textsuperscript{3}-Human can achieve high-quality decoupled reconstruction of the human body wearing different clothing, and can be directly applied to clothing transfer and animation.

\end{abstract}    
\section{Introduction}
\label{sec:intro}

Clothed human body reconstruction has long been a research focus in the fields of graphics and computer vision, and has a wide range of applications in many fields such as virtual reality, augmented reality, holographic communication, film production, and game development. 
Compared to film-level reconstructions requiring numerous cameras and artists for modeling, reconstructing high-quality clothed human bodies from monocular videos holds more practical value for general users. 
In scenarios such as telepresence and virtual try-ons, the 3D avatars used should be easily accessible, visually realistic, and easy to edit, including modifications to clothing and posture. 
Therefore, how to reconstruct high-fidelity, decoupled representations of clothed human bodies using monocular video remains a long-standing research problem. 
Through decoupled reconstruction, clothing can be separated from the human body, enabling efficient adjustment and editing of different clothing styles, postures, and body shapes.
This decoupling not only enhances the flexibility and practicality of 3D reconstruction, but also improves the realism of detailed features, providing greater potential for personalization and dynamic interaction of virtual characters.

We aim to develop a method to decouple and reconstruct the clothed human body from monocular videos. However, this is a very challenging task because 1) monocular videos only provide single-view 2D image information and lack direct 3D depth perception, 2) real captured videos contain various clothing styles, irregular textures, and complex human poses, 3) body parts occluded by clothes are not visible in the input video, which also poses a great challenge to reconstruction. Existing methods can be divided into explicit expression methods and implicit expression methods. Among them, explicit expression methods usually rely on pre-acquired templates. Some methods ~\cite{deepcap,livecap,Xu2018MonoPerfCap} use scanners, while other methods ~\cite{alldieck2018videoavatar, li2023tcsvt, pramishp2024iHuman} rely on parametric models ~\cite{loper2015smpl, jiang2020bcnet}, and the reconstruction quality mainly depends on the representation ability of the model. Implicit representation methods ~\cite{Feng2022scarf, peng2023intrinsicngp, jiang2022neuman, jiang2022selfrecon, guo2023vid2avatar} use NeRF~\cite{mildenhall2020nerf} or SDF~\cite{park2019deepsdf} to model the clothed human body, but they usually produce an inseparable whole body or exhibit average geometric quality.

We propose a decoupled human body reconstruction scheme, named $\text{D}^3\text{-Human}$ (Dynamically Disentangled Digital Human), which combines explicit and implicit representations to address the main challenges of template generation and the dynamic deformations.
In decoupled human body reconstruction, generating clothing templates is particularly challenging.
Traditional methods, such as those based on parametric models ~\cite{li2023tcsvt} or feature lines ~\cite{qiu2023RECMV}, rely heavily on priors, limiting the types of clothing they can represent.
Although implicit unsigned distance field (UDF) representations ~\cite{guillard2022udf, long2022neuraludf, Liu23NeUDF, Chen2024NeuralABC} provide some solutions, they perform poorly when single-view supervision is limited (see experiments~\ref{sec:udf}).
For the visible region, inspired by GShell ~\cite{liu2024gshell} and DMTet ~\cite{shen2021dmtet}, we define an optimizable human manifold signed distance field (hmSDF) on the non-decoupled clothed human surface to separate clothing from the body.
To the best of our knowledge, this is the first method that can reconstruct clothing geometry from monocular dynamic human videos without any 3D clothing priors, using only easily obtainable 2D human parsing segmentation~\cite{ravi2024sam2}.
For the invisible regions of the body, we adopt the explicit representation of the corresponding regions of the SMPL ~\cite{loper2015smpl} model to ensure the plausibility of the body shape and seamless integration with the visible region.
This approach enables detailed and decoupled reconstruction of clothed human bodies.

We reconstruct different human bodies wearing various outfits based on monocular videos, to demonstrate the capabilities of our method.
Compared to existing methods that require over a day~\cite{jiang2022selfrecon, qiu2023RECMV, wen2024gomavatar} for reconstruction, $\text{D}^3\text{-Human}$ reconstructs a decoupled template of clothing and body in a much shorter time (about 20 minutes) and completes the full sequence within several hours, achieving competitive reconstruction accuracy.
Furthermore, we demonstrate examples of applications in animation production and clothing transfer to showcase the wide-ranging applicability of the decoupled representation.
In summary, the contributions of this paper include the following aspects:
\begin{itemize}
    \item A hybrid reconstruction method combining explicit and implicit representations is proposed, which is capable of reconstructing high-quality, decoupled clothing and human body from monocular video. 
    \item For the clothed human body in visible areas, we introduce a novel representation, hmSDF, which can accurately segment 3D clothing and body through easily obtainable 2D human parsing, without any 3D clothing priors.
    \item The reconstructed decoupled clothing and human body can be easily applied to animation production and clothing transfer applications, offering realistic and detailed geometric quality.
\end{itemize}

\section{Related Work}
\label{sec:formatting}

\textbf{Decoupled Representation of Clothed Human Body.}
Most methods reconstruct the clothed human body geometry as a whole, including representations like mesh~\cite{ThiemoAlldieck2019Tex2ShapeDF, alldieck2018videoavatar}, point clouds~\cite{POP:ICCV:2021, zhang2023closet}, SDF~\cite{chen2022gdna, palafox2021npms, jiang2022selfrecon}, and occupancy~\cite{saito2019pifu, saito2020pifuhd}. 
These reconstruction methods maintain good detail in visible areas but are inconvenient for applications like animation and clothing transfer. 
A more effective approach is to independently represent clothing and the body, modeling them as decoupled, layered representations. 
GALA~\cite{kim2024gala} and ClothCap~\cite{Sig2017ClothCap} utilize 3D segmentation to obtain separate clothing and body meshes from 3D and 4D scans, respectively, a method constrained by the high cost of acquiring scan data. 
Neural-ABC~\cite{Chen2024NeuralABC} constructs a decoupled human body and clothing parametric model based on UDF (unsigned distance function) representation, but its reconstruction of details is limited. 
SCARF~\cite{Feng2022scarf} reconstructs a mixed representation of a clothed human body from video sequences, but the NeRF~\cite{mildenhall2020nerf} representation of clothing is limited in geometric effect.
Some other methods focus on reconstructing clothing~\cite{jiang2020bcnet, de2023drapenet, corona2021smplicit, 
 Zhu_2022_CVPR, li2023isp}, which allows decoupling by using SMPL~\cite{loper2015smpl} as the underlying body; however, the body often lacks detail in these approaches.

\textbf{Reconstruction from a Single-View Image.}
Traditional methods reconstruct by fitting a parametric human body model to a single-view image through optimization or regression within a parametric space~\cite{lin2021end-to-end, li2022cliff, kanazawa2018end}. 
The effectiveness of the fit largely depends on the representational capacity of the parametric model. 
Models like SMPL~\cite{loper2015smpl} and SCAPE~\cite{Anguelov05scape:shape} can reconstruct the underlying body without clothes.
Some approaches~\cite{jiang2020bcnet, corona2021smplicit, de2023drapenet} can reconstruct clothing and use the SMPL model to represent the underlying body, achieving a complete reconstruction of the clothed human body. 
There are also models that simultaneously represent both clothing and the underlying body, including methods that model both uniformly~\cite{palafox2021npms,chen2022gdna} and hierarchically~\cite{Chen2024NeuralABC}. 
Parametric-based reconstruction methods can easily produce plausible clothed human figures but generally lack detail.
Methods that target single images can reconstruct videos frame by frame; however, they often fail to ensure inter-frame consistency, such as lacking frame-to-frame coherence, or may lead to unsmooth results.

\textbf{Reconstruction from Monocular Video.}
3D clothed human body reconstruction from video inputs typically relies on motion and deformation cues to recover deformable 3D surfaces. 
Early works acquired actor-specific rigged templates~\cite{Xu2018MonoPerfCap, deepcap, livecap} or used a parametric model as a prior~\cite{alldieck2018videoavatar}. 
Many efforts~\cite{Feng2022scarf, peng2023intrinsicngp, jiang2022neuman, peng2021neural, wen2024gomavatar, pramishp2024iHuman}, based on NeRF~\cite{mildenhall2020nerf} and 3DGS~\cite{kerbl3Dgaussians}, reconstruct animatable human avatars from dynamic videos, primarily focusing on rendering effects, but the quality of geometric reconstruction is not ideal.  
~\cite{jiang2022selfrecon, tan2024dressreconfreeform4dhuman, guo2023vid2avatar} reconstructed high-quality clothed human body geometries, but the clothing and underlying body are not separable.
DGarment~\cite{li2023tcsvt} and REC-MV~\cite{qiu2023RECMV} reconstructed dynamic clothing, excluding the body. 
Our method can reconstruct decoupled clothing and the underlying body while ensuring geometric quality.

\section{Method}

Given a monocular video containing $N$ frames, which depicts a person in motion wearing clothing, as $\{I_t \mid t = 1, \ldots, N\}$.
\(\text{D}^3\text{-Human}\) aims to reconstruct high-fidelity, decoupled, and spatio-temporally continuous clothing and underlying body meshes \(\{G_t \mid t = 1, \ldots, N\}\) without using 3D clothing template priors.
With the goal of achieving maximum realism, the observed areas captured by the video, 
such as the exposed head and clothing, should be reconstructed with great detail; 
areas obscured, such as body parts covered by clothing or self-occluded by the body, should be reconstructed as plausibly as possible.

\begin{figure*}[ht]
\centering
\includegraphics[scale=0.145]{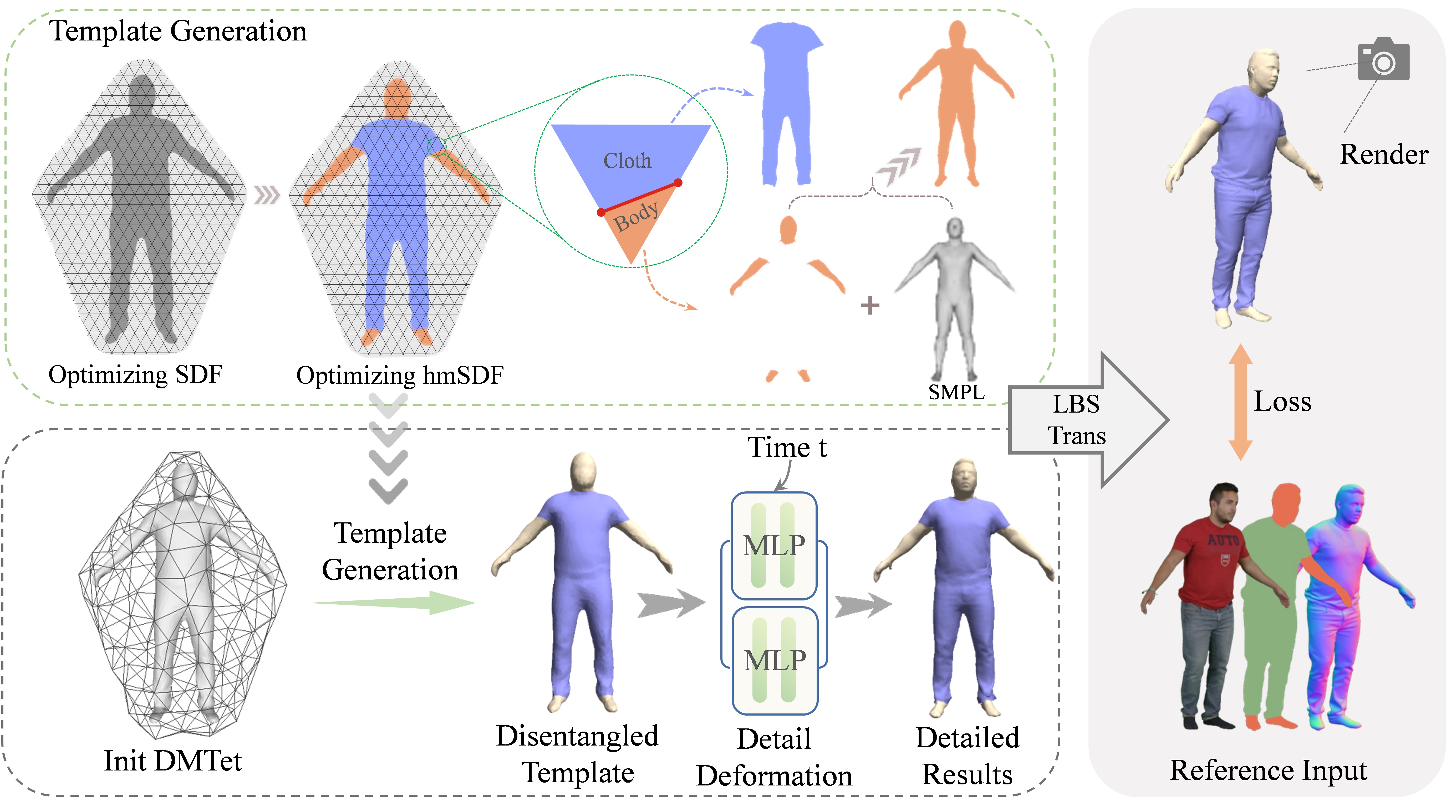}
\vspace{-1mm}
\caption{Overview of $\text{D}^{3}$-Human. 
The optimization process is divided into two steps: template generation and detailed deformation.  
The object is initialized as a DMTet~\cite{shen2021dmtet} representation, and is optimized to form a complete clothed human. 
An optimizable HmSDF function separates the clothing and body regions, with missing parts filled by SMPL. 
After generating the disentangled template, we use two MLPs to model detailed deformations for each frame of the body and clothing meshes separately. 
Finally, the meshes are transformed to the observed space using a forward LBS deformation, supervised by images, normal maps, and parsing masks with a differentiable renderer.
}
\label{fig_pp1}
\end{figure*}

To achieve these objectives, 
We combine the flexibility of implicit representations with the robustness and rapid rendering capabilities of explicit representations to achieve the best results. 
Notably, the surface of a watertight clothed human can be segmented into clothing and body parts by closed curves. 
Therefore, we first utilize image and human parsing segmentation sequence information to reconstruct the separated clothing and body meshes in the visible areas.
Then, with the help of SMPL~\cite{loper2015smpl}, we complete the invisible body areas and generate decoupled clothing and body templates. 
Finally, normal information is additionally used to jointly optimize the clothing and body to enhance details.
Figure~\ref{fig_pp1} illustrates the overall pipeline of our approach. 

\subsection{HmSDF Representation}

In this section, we define the clothing template \(G_c\) and the body template \(G_b\) in canonical space. 
Considering that the body parts obscured by clothing are invisible in the video, we further divide the clothing and body into the visible body \(S_b\), the invisible body \(U_b\), and the visible clothing \(S_c\). 
The visible body \(S_b\) is obtained via segmentation after reconstructing the whole clothed body, while the invisible body \(M_b\) is completed using the SMPL model. 
The visible and invisible bodies are merged to form the body template \(G_b\).

The visible clothing and body are represented by a hybrid representations\cite{shen2021dmtet} that combines a tetrahedral mesh grid \( (V_T, T) \) and a neural implicit signed distance function \( s_{\eta}(x) \), 
where \( x \in V_T \) and \( s_{\eta}(x) \) is a neural network with learnable weights \( \eta \). 
The surface of \( S_b \cup S_c \) can be represented by \( S_{\eta} = \{x \in \mathbb{R}^3 \mid s_{\eta}(x; \eta) = 0\} \). 
The mesh can be extracted using methods following GShell~\cite{liu2024gshell}.

Since the reconstructed watertight clothed human body \( S_{\eta} = S_b \cup S_c \), 
% \( S_{\eta} \) consists solely of \( S_b \) and \( S_c \), 
we define a continuous and differentiable mapping \( \nu : S_{\eta} \to \mathbb{R} \) on \( S_{\eta} \) to characterize whether a point belongs to \( S_b \) or \( S_c \):
\[
\nu(x) = 
\begin{cases} 
< 0, & \forall x \in \text{Interior}(S_b),\\
= 0, & \forall x \in \lambda, \\
> 0, & \forall x \in \text{Interior}(S_c), \\
\end{cases}
\]
where \( \lambda \) represents the boundary line between \( S_b \) and \( S_c \).
We denote \( \nu \) as the human manifold signed distance fields, termed hmSDF.
This differs from the definition of mSDF in GShell, which only considers points located inside the open surface. 
In contrast, our approach considers points on both sides of the hmSDF (the visible clothing and body).

\subsection{Region Aggregation.}
\label{Segmentation}

\begin{figure}[ht]
  \centering
  \includegraphics[scale=0.25]{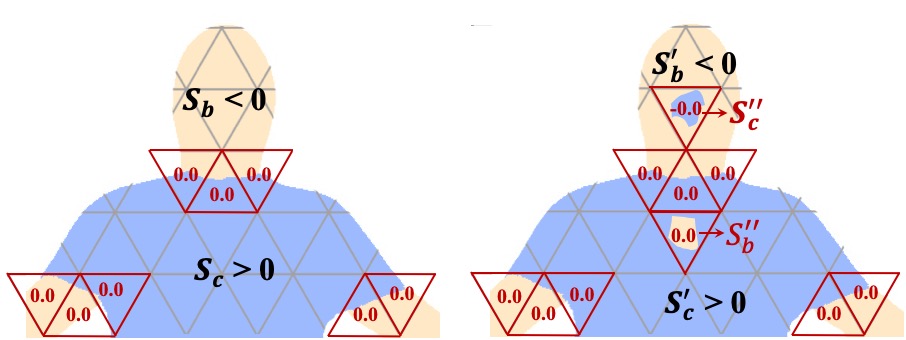}
  \vspace{2mm}
  \begin{picture}(0,0)
    \put(-210,-11){Correct Segmentation}
    \put(-110,-11){Inaccurate Segmentation}
  \end{picture}
  \caption{Schematic of region aggregation.
  For the correct segmentation results, \( S_b \) and \( S_c \) correctly segment the body and the cloth. For inaccurate segmentation results, \( S_c^{''} \) should merge with \( S_b^{'} \), and \( S_b^{''} \) should merge with \( S_c^{'} \).
  }
  \label{agg}
\end{figure}

Ideally, an optimized hmSDF function \( \nu \) should accurately segment the visible clothing \( S_c \) and body \( S_b \). However, due to inaccuracies in the human parsing mask of each frame and inconsistencies between frames, inaccuracies may occur in the boundary line \( \lambda \)
, as illustrated in Figure~\ref{agg}.
When inaccuracies occur in the neighborhood of \( \nu(x) = 0 \), it causes the segmented regions \( S_b^{'} \) and \( S_c^{'} \) to contain holes and creates undesired segmented regions \( S_b'' \) and \( S_c'' \).
The small fragments contained in \( S_b'' \) and \( S_c'' \) are inevitably disconnected from other subgraphs. 

We obtain the number of connected components for each category from the input image. Since the vertices of incorrectly classified connected components are usually fewer, the category of each subgraph, \( S_b' \), \( S_c' \), \( S_b'' \), and \( S_c'' \), can be determined by calculating the number of vertices in each connected component using the depth-first search method.
The correct \( S_b \) and \( S_c \) can be obtained by aggregation as follows:
\begin{equation}
  S_b = \text{merge}(S_b', S_c''),
  \label{eq:merge1}
\end{equation}
\begin{equation}
  S_c = \text{merge}(S_c', S_b'').
  \label{eq:merge2}
\end{equation}

%---------------------------------------
We demonstrate the process of aggregation of \( S_b  \) and \( S_c \) in Algorithm~\ref{algorithm: close_hole}.

\begin{algorithm}[!htb]
	\renewcommand{\algorithmicrequire}{\textbf{Input:}}
	\renewcommand{\algorithmicensure}{\textbf{Output:}}
	\caption{Region Aggregation Algorithm}
	\label{algorithm: close_hole}
    	% \begin{algorithmic}[1]
    		
                \textbf{Input:}  The initialization segmentation \( S_b^0 \) and \( S_c^0 \) directly obtained from \( \lambda \);
                  The correct number of subgraphs for \( S_b \) and \( S_c \) are \( \alpha_1 \) and \( \alpha_2 \), respectively.
            
                     \hspace{0.5cm} Calculate all connected subsets \( Q_b \) of \( S_b^0 \); 
                     
                     \hspace{0.5cm} Calculate all connected subsets \( Q_c \) of \( S_c^0 \);
                     
                     \hspace{0.5cm} Sort \( Q_b \) based on the number of vertices;
                     
                     \hspace{0.5cm} Sort \( Q_c \) based on the number of vertices;
                     
                     \hspace{0.5cm} Extract \( S_b' \) and \( S_b'' \) based on \( \alpha_1 \) from \( Q_b \);
                     
                     \hspace{0.5cm} Extract \( S_c' \) and \( S_c'' \) based on \( \alpha_2 \) from \( Q_c \);
                     
                     \hspace{0.5cm} Obtain \( S_b \) by Equation~\ref{eq:merge1} and filter out duplicate points.
                     
                     \hspace{0.5cm} Obtain \( S_c \) by Equation~\ref{eq:merge2} and filter out duplicate points.
                
                \textbf{return~}\( S_b \) and \( S_c \) without holes and fragments.
            % \end{algorithmic}
\end{algorithm}

\subsection{Deformation Fields}

Similar to previous methods~\cite{jiang2022selfrecon,qiu2023RECMV}, we use the SMPL-based Linear Blend Skinning (LBS) method to model large deformations based on skeletal movements, and employ non-rigid deformation fields to model subtle deformations. 
However, a key difference is that clothing and body follow different motion rules. 
Therefore, we use two separate non-rigid deformation fields to model the non-rigid deformations of clothing and body respectively.

\textbf{Non-rigid Deformation.}
Due to its limited degrees of freedom, LBS deformation can only model large deformations and is unable to represent smaller details, such as the folds of clothing. 
Therefore, for detail deformation, 
we use two MLPs to model the non-rigid deformations of clothing and the human body.
\( D \) is the MLP for non-rigid deformations:
\[
x^t = D(x, h^t, E(x); \phi),
\]
where \( x \) is the point in canonical space, \( x^t \) is the point after deformation in frame \( t \), \( h^t \) is the latent code corresponding to frame \( t \), and \( \phi \) are the network parameters that need to be optimized. 
For clothing and the body, the networks and parameters are independently separate.

\textbf{LBS Deformation.}
Linear Blend Skinning (LBS) deformation models the transformation from canonical space to observed space based on skeletal deformation. 
Given the SMPL shape parameters \(\beta\) and pose parameters \(\theta_t\), the LBS deformation \(W\) can be written as:
\[
G^{'}(\beta, \theta_t) = W(D(x), \beta, \theta_t, \mathcal{W}(x)),
\]
where \(D(x)\) represents the non-rigid deformation of clothing and body, and \(\mathcal{W}(x)\) is the method for computing the skinning weights of \(x\) based on SMPL. 
We refer to some clothing simulation methods~\cite{grigorev2022hood, santesteban2022snug} to calculate the skinning weights.
For both clothing and body, we use a shared skinning deformation model.

\subsection{Occlusion-Aware Differentiable Rendering}
Differentiable rendering is used to render the geometry in the observed space to 2D, which allows for the calculation of loss with 2D supervision.
Following some differentiable rendering methods~\cite{liu2024gshell, hasselgren2022nvdiffrecmc,Munkberg_2022_CVPR}, we utilize a differentiable rasterization approach to render the mesh. 
Compared to volumetric rendering methods, rasterization-based rendering enables differentiable rendering for explicit meshes and offers better time and memory efficiency.

\begin{figure}[ht]
\centering
\includegraphics[scale=0.2]{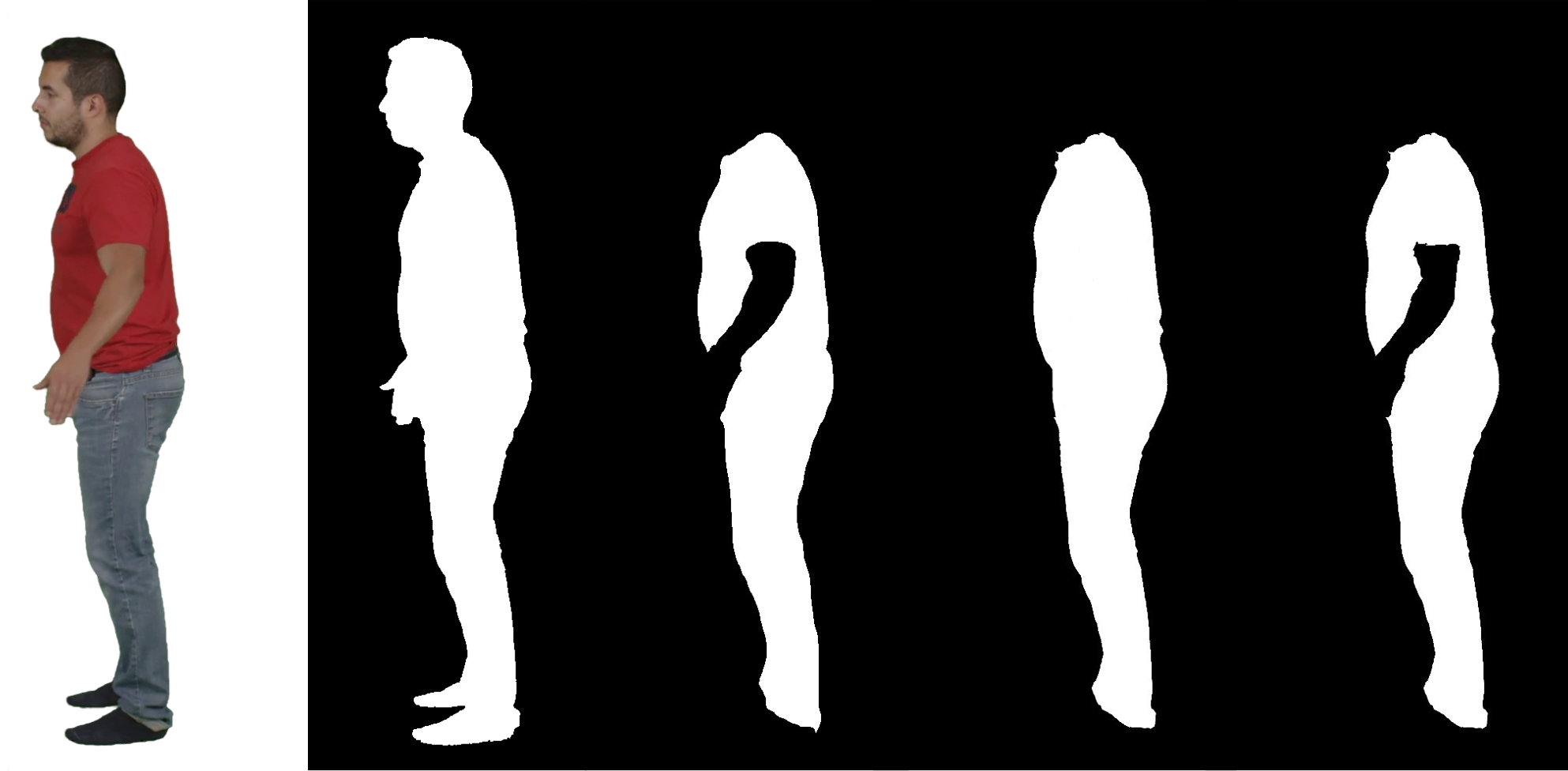}
\caption{Occlusion display of the mask. 
From left to right: the color image of the captured clothed human, the complete clothed body mask obtained from SAM2~\cite{ravi2024sam2}, the clothing mask obtained from SAM2, the mask obtained by rendering only the clothing mesh, and the mask of the effective clothing area after rendering the complete clothed body mesh.
}
\label{mask_occ}
\end{figure}

For clothed bodies, occlusion may occur between the body and clothing, which can lead to occlusion when rendering the visible areas of the clothing from the same viewpoint. 
Therefore, only rendering the clothing mesh to obtain a clothing mask may produce results inconsistent with the supervision signals, as illustrated in Figure~\ref{mask_occ}.
To solve this problem, we label the faces of both the clothing and body, rendering them simultaneously. 
We then use rasterization to generate occlusion-aware 2D labels, where the effective area of the body label is the body mask \(M_b\), 
and the effective area of the clothing label is the clothing mask \(M_c\).

\subsection{Training}

\subsubsection{Training Strategy}
Our method consists of two stages: template generation and detail deformation optimization. In the template generation stage, we leverage hmSDF directly learn the clothing template from mask supervision, without relying on 3D clothing priors. In this stage, deformation is achieved solely through LBS, and parameters are optimized through RGB Loss, Mask Loss, Eikonal Loss, Encourage Hole Opening, and Regularize Holes. 
In the detail deformation stage, we introduce an additional Perceptual Normal Loss as a reconstruction term, while regularization is applied through Collision Penalty and Geometry Regularization to optimize non-rigid deformation.

\subsubsection{Reconstruction Loss}

We minimize the difference between the rendered result and the input image through the following objective: 

\textbf{RGB Loss.}
The $L_1$ is calculated between the rendered RGB image and the supervision image. 
We calculate for all valid pixels \( P \) as following,
\begin{equation}
\begin{aligned}
\text{L}_{\text{color}} = \frac{1}{|P|} \sum_{p \in P} \Big( & \mathds{1}_{\text{b}}(\bm{p}) \cdot \text{L}_{\text{mse}}(I_b, \hat{I_b}) \\
+ & \mathds{1}_{\text{c}} (\bm{p}) \cdot \text{L}_{\text{mse}}(I_c, \hat{I_c}) \Big),
\end{aligned}
\end{equation}
where $\mathds{1}(\bm{p})$ indicates the category to which the current pixel point $\bm{p}$ belongs. When $\mathds{1}_{\text{b}}(\bm{p})$ is true, pixel $\bm{p}$ belongs to the body; when $\mathds{1}_{\text{c}}(\bm{p})$ is true, pixel $\bm{p}$ belongs to the cloth. Both $\mathds{1}_{\text{b}}(\bm{p})$ and $\mathds{1}_{\text{c}}(\bm{p})$ can be true simultaneously, or only one may be true, depending on the mask used for supervision.
\(I_b\) is the rendered RGB image of the body, \(I_c\) is the rendered RGB image of the clothing, \(\hat{I_b}\) is the ground truth RGB image of the body, and \(\hat{I_c}\) is the ground truth RGB image of the clothing.

\textbf{Mask Loss.}
Although irrelevant backgrounds have already been removed in the RGB image supervision, the independently added mask loss can further constrain the accuracy of the edges, as,
\begin{equation}
\begin{aligned}
\text{L}_{\text{mask}} &= \frac{1}{|P|} \sum_{p \in P} \Big( \mathds{1}_{\text{b}}(\bm{p}) \cdot \text{L}_{\text{mse}}(M_b, \hat{M_b}) \\
&\quad +  \mathds{1}_{\text{c}} (\bm{p}) \cdot \text{L}_{\text{mse}}(M_c, \hat{M_c}) \Big),
\end{aligned}
\end{equation}
where 
\(\hat{M_b}\) is the ground truth mask of the body, and \(\hat{M_c}\) is the ground truth mask of the clothing.

\textbf{Perceptual Normal Loss.}
We obtain the normals of the image through Sapiens~\cite{khirodkar2024sapiens} as ground truth, to leverage the prior information trained on large-scale human body data.
Rendered normals and supervision normals need to be normalized and aligned to the observation space. 
We use the perceptual loss ~\cite{johnson2016perceptual, hong2021headnerf} to further enhance the effectiveness of the rendered normals.
\begin{equation}
\begin{aligned}
\text{L}_{\text{per}} = \sum_i \|\phi_i(\mathcal{N}) - \phi_i(\hat{\mathcal{N}})\|^2,
\end{aligned}
\end{equation}
where \(\mathcal{N}\) is the rendered normal, and \(\hat{\mathcal{N}}\) is the ground truth normal, and $\phi_i(*)$ denotes the activation of the \(i\)-th layer in the MobileNetV2 network ~\cite{sandler2018mobilenetv2}. 

\subsubsection{Regularization Term}

\textbf{Eikonal Loss.} To ensure a reasonable signed distance field, we add an Eikonal term~\cite{icml2020_2086} to the gradient \( g \) of the SDF value at each tetrahedral vertex when optimizing the SDF:
\begin{equation}
\begin{aligned}
\text{L}_{\text{eik}} = \sum_{u \in V_T} (\|g_u\|_2 - 1)^2.
\end{aligned}
\end{equation}

\textbf{Encourage Hole Opening.}
With limited viewpoints, it is necessary to identify the opening positions using only image information. 
We encourage hmSDF openings by adopting a regularization term similar with
~\cite{liu2024gshell} as,
\begin{equation}
\begin{aligned}
\text{L}_{\text{hole}} = \sum_{u:\nu(u) \geq 0} \text{L}_{\text{huber}}(\nu(u)).
\end{aligned}
\end{equation}

\textbf{Regularize Holes.}
To avoid excessively large openings, we impose constraints on all points that are visible from the current viewpoint as
\begin{equation}
\begin{aligned}
\text{L}_{\text{reg-hole}} = \sum_{u:\nu(u) = 0} \text{L}_{\text{huber}}(\nu(u)  - {\epsilon}_{1}),
\end{aligned}
\end{equation}
where $\epsilon_{1}$ is a positive scalar.

\textbf{Collision Penalty.} 
This ensures that the garment does not penetrate the underlying body, inspired by ~\cite{grigorev2022hood, santesteban2022snug}. We implement it as
\begin{equation}
\begin{aligned}
\text{L}_{\text{collision}} = \sum_{\text{vertices}} k_{\text{collision}} \max({\epsilon}_{2} - d(x), 0)^3.
\end{aligned}
\end{equation}
In particular, when the distance between the two layers is too close, rendering can produce computational errors, so the value of \(\epsilon_2\) is set to 0.005.

\textbf{Geometry Regularization.} 
To ensure that the optimization is constrained, we encourage generating smooth deformed results.
Inspired by \citet{worchel2022nds}, we add normal consistency term \(\text{L}_{\text{n\_consist}}\) and Laplacian term \(\text{L}_{\text{laplacian}}\).

\section{Experiments}

We conduct qualitative and quantitative experiments to demonstrate the effectiveness of $\text{D}^3\text{-Human}$. 
For qualitative experiments, we use subjects from PeopleSnapshot~\cite{alldieck2018videoavatar} and SelfRecon~\cite{jiang2022selfrecon}. 
For quantitative experiments, we use the synthetic dataset constructed by SelfRecon. It provides accurate ground truth for the meshes.
Additionally, we perform ablation studies on the discussion of UDF and hmSDF, as well as the effectiveness of the perceptual normal loss, and demonstrate applications in clothing transfer and physics-based animation production.

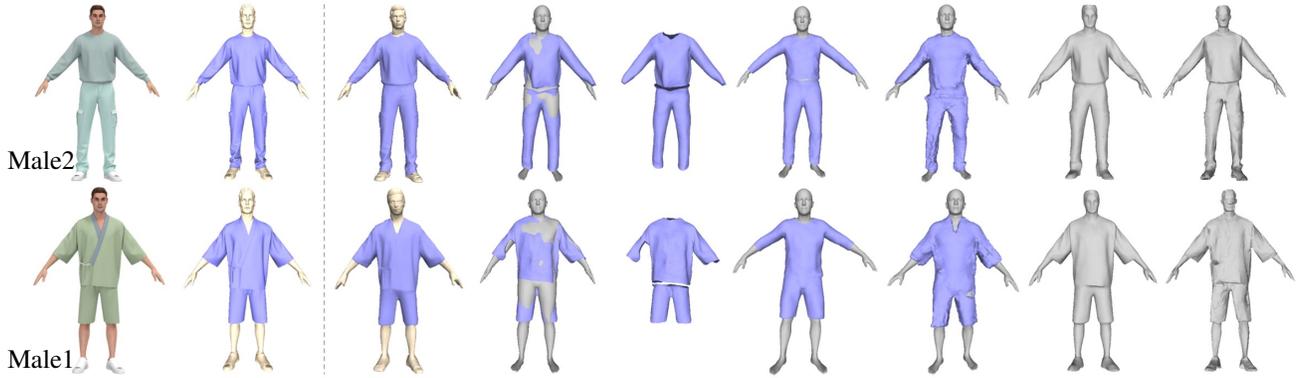
\begin{figure*}[ht]
\centering
 \includegraphics[scale=0.43]{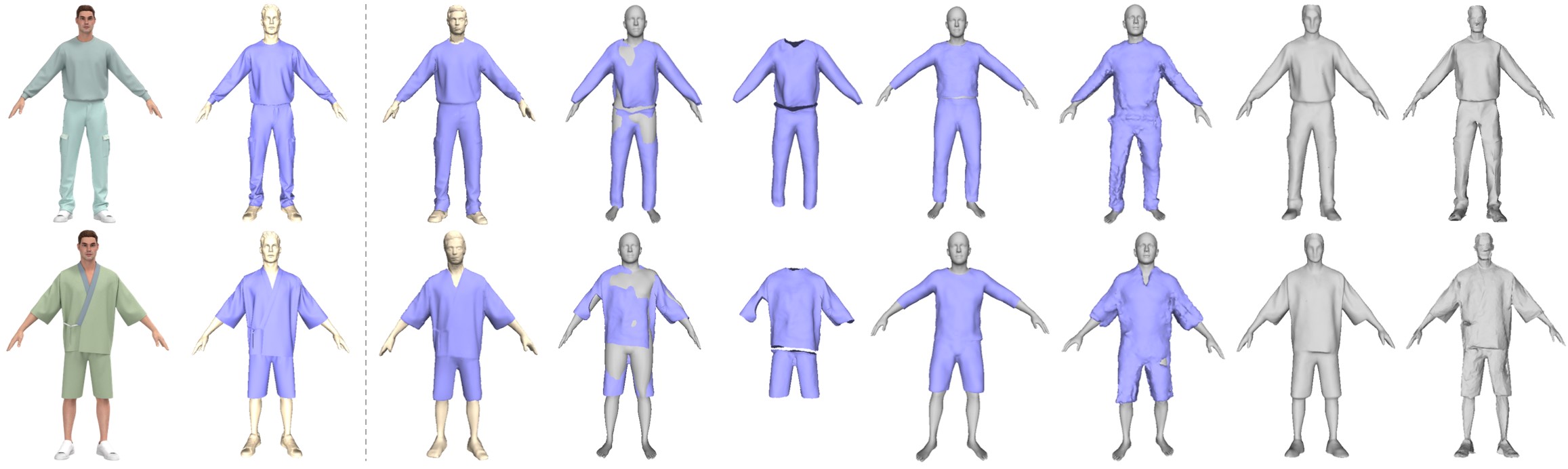}
\vspace{1mm}
\begin{picture}(0,0)
    \put(-490,-9){Reference Images}
    \put(-490,80){Male2}
    \put(-490,5){Male1}
    \put(-145,-9){DELTA}
    \put(-405,-9){GT}
    \put(-355,-9){Ours}
    \put(-330,-9){REC-MV w/ SMPL}
    \put(-245,-9){REC-MV}
    \put(-200,-9){BCNet}
    \put(-145,-9){DELTA}
    \put(-100,-9){SelfRecon}
    \put(-50,-9){GoMAvatar}
\end{picture}
\caption{
Quantitative  comparison of the proposed method with REC-MV~\cite{qiu2023RECMV}, BCNet~\cite{jiang2020bcnet}, DELTA~\cite{Feng2023DELTA}, SelfRecon~\cite{jiang2022selfrecon}, and GoMAvatar~\cite{wen2024gomavatar}.
We use purple to visualize clothing that can be decoupled from the body.
For REC-MV and BCNet, SMPL~\cite{loper2015smpl} was added as the body to show the complete reconstruction of the clothed human.
}

\label{compare_num}
\end{figure*}

\begin{table*}[ht]
\centering
\caption{Quantitative comparison across four synthetic sequences.
We report the Chamfer Distance (CD) between the reconstructed surfaces (cm) and ground truth. 
For REC-MV, BCNet, DELTA and our method, we report the CD for clothing, body, and the full clothed human body, respectively. 
For SelfRecon and GoMavatar, we only report the CD for the full clothed human body. The unit is $e^{-3}$.
We highlight the \best{best} value and the \second{second best} value.
We show Male1 and Male2 in Figure~\ref{compare_num}, and Female1 and Female3 in the supplementary materials.
}
\vspace{-2mm}
\label{tab1}
\resizebox{\textwidth}{!}{
    \begin{tabular}{lccc|ccc|ccc|ccc}
    \toprule
    \multirow{2}{*}{Method} & \multicolumn{3}{c|}{Female1} & \multicolumn{3}{c|}{Female3} & \multicolumn{3}{c|}{Male1} & \multicolumn{3}{c}{Male2} \\
    & Clothing & Body & All & Clothing & Body & All & Clothing & Body & All & Clothing & Body & All \\
    \midrule
    REC-MV~\cite{qiu2023RECMV} & \second{1.416} & 1.789 & \second{1.148} & \second{0.930} & 2.082 & 1.461 & \second{0.614} & 1.945 & \second{0.619} & \second{0.693} & 1.201 & \second{0.616} \\
    BCNet~\cite{jiang2020bcnet} & 1.685 & 10.252 & 5.561 & 4.571 & 10.112 & 5.681 & 2.589 & 6.236 & 4.802 & 2.007 & 4.109 & 2.853 \\
    DELTA~\cite{Feng2023DELTA} & 2.177 & \second{0.973} & 1.388 & 2.173 & \second{0.820} & \second{0.915} & 1.327 & \second{1.498} & 1.702 & 1.884 & \second{1.073} & 1.132 \\
    SelfRecon~\cite{jiang2022selfrecon} & - & - & 3.420 & - & - & 2.249 & - & - & 1.310 & - & - & 1.454 \\
    GoMavatar~\cite{wen2024gomavatar} & - & - & 7.319 & - & - & 5.058 & - & - & 2.382 & - & - & 3.163 \\
    Ours & \best{\textbf{1.065}} & \best{\textbf{0.966}} & \best{\textbf{0.959}} & \best{\textbf{1.109}} & \best{\textbf{0.742}} & \best{\textbf{0.636}} & \best{\textbf{0.478}} & \best{\textbf{0.321}} & \best{\textbf{0.270}} & \best{\textbf{0.355}} & 
    \best{\textbf{0.325}} & \best{\textbf{0.279}} \\
    \bottomrule
    \end{tabular}%
}
\end{table*}

\begin{figure}[ht]
  \centering
  \includegraphics[scale=0.41]{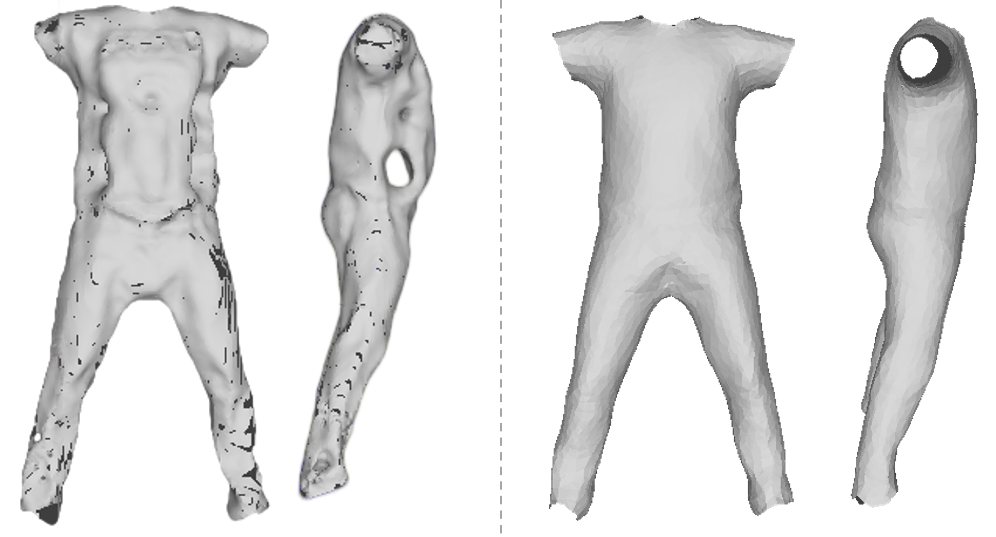}
  \begin{picture}(0,0)
    \put(-180,-7){Implicit UDF}
    \put(-55,-7){HmSDF}
  \end{picture}
  \caption{Ablation study on clothing reconstruction using implicit UDF~\cite{long2022neuraludf} with deformation field~\cite{jiang2022instantavatar} and hmSDF, applied to the male-3-casual data from the PeopleSnapshot~\cite{alldieck2018videoavatar} dataset.
  }
  \label{ablation_udf}
\end{figure}

\begin{figure}[ht]
  \centering
  \includegraphics[scale=0.41]{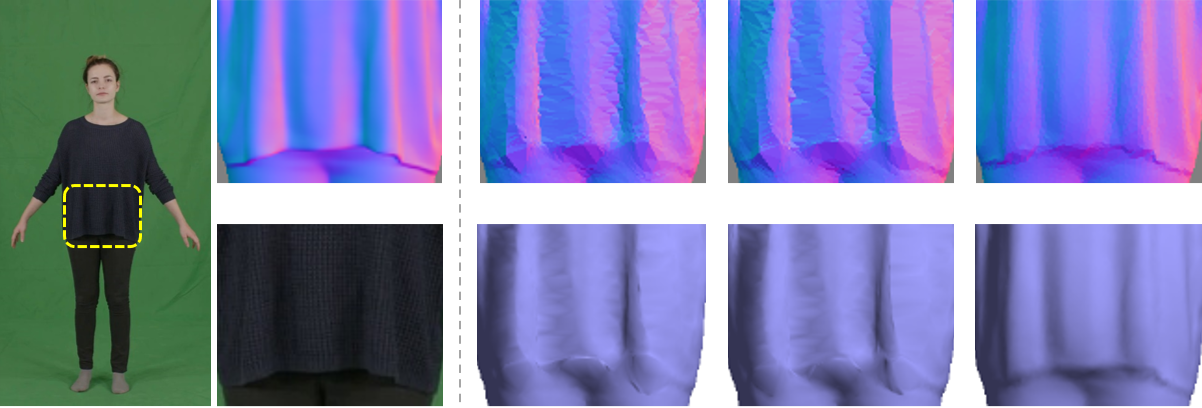}
  \begin{picture}(0,0)
    \put(-103,-4){Reference Input}
    \put(-10,-4){COS}
    \put(38,-4){MSE}
    \put(75,-4){Perception}
  \end{picture}
  \caption{Ablation study on normal loss.
  The reference input includes normals and images. 
  The top row shows the rendered normals, and the bottom row shows the rendered meshes.
  }
  \label{ablation_normal}
\end{figure}

\begin{figure*}[ht]
\centering
\includegraphics[scale=0.54]{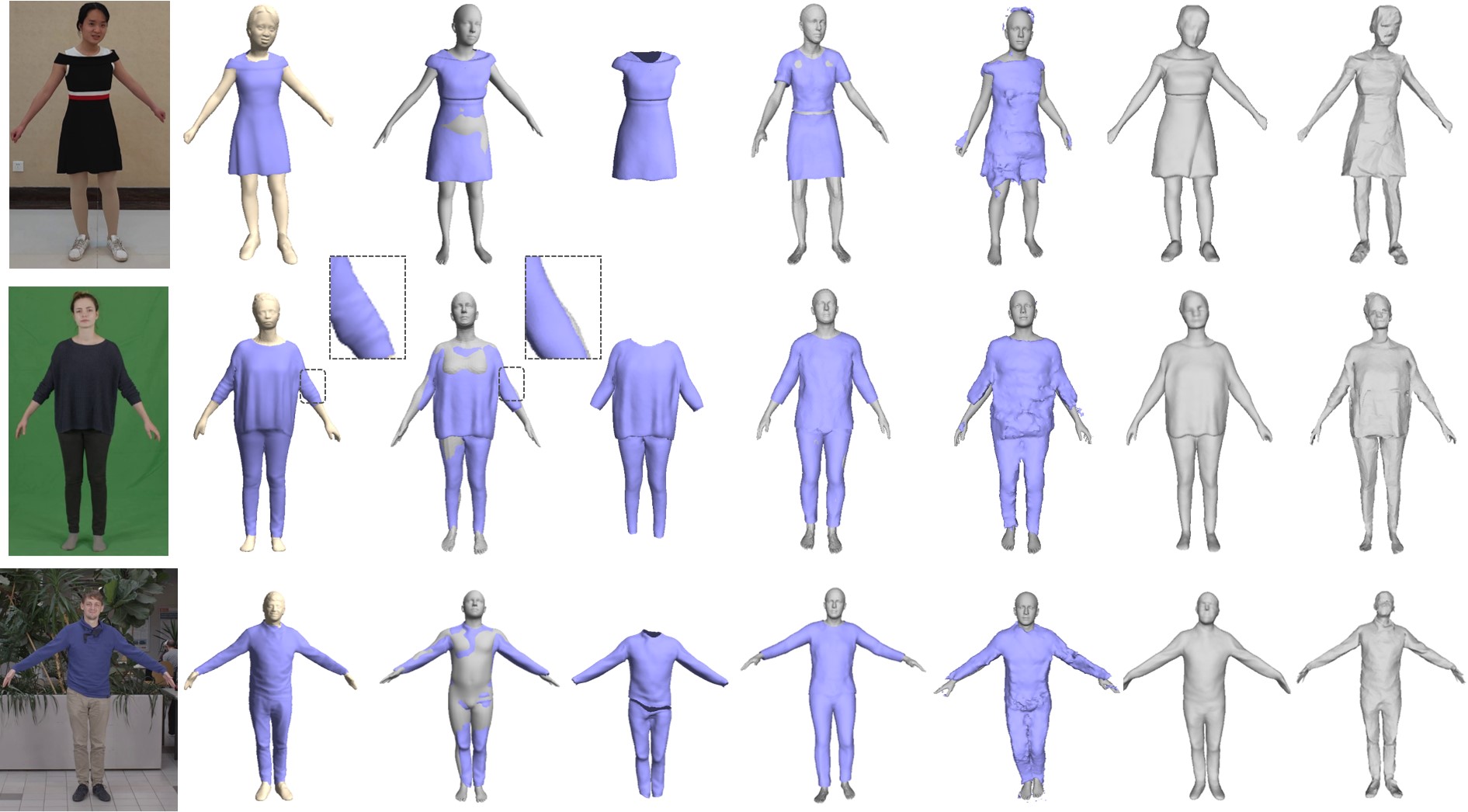}
\vspace{3mm}
\begin{picture}(0,0)
    \put(-495,-12){Reference Input}
    \put(-413,-12){Ours}
    \put(-380,-12){REC-MV w/ SMPL}
    \put(-290,-12){REC-MV}
    \put(-225,-12){BCNet}
    \put(-165,-12){DELTA}
    \put(-113,-12){SelfRecon}
    \put(-50,-12){GoMAvatar}
\end{picture}
\caption{
Qualitative comparison.
Comparison of our method with other methods on real image sequences.
}
\label{compare2}
\end{figure*}

\begin{figure}[ht]
  \centering
  \includegraphics[scale=0.35]{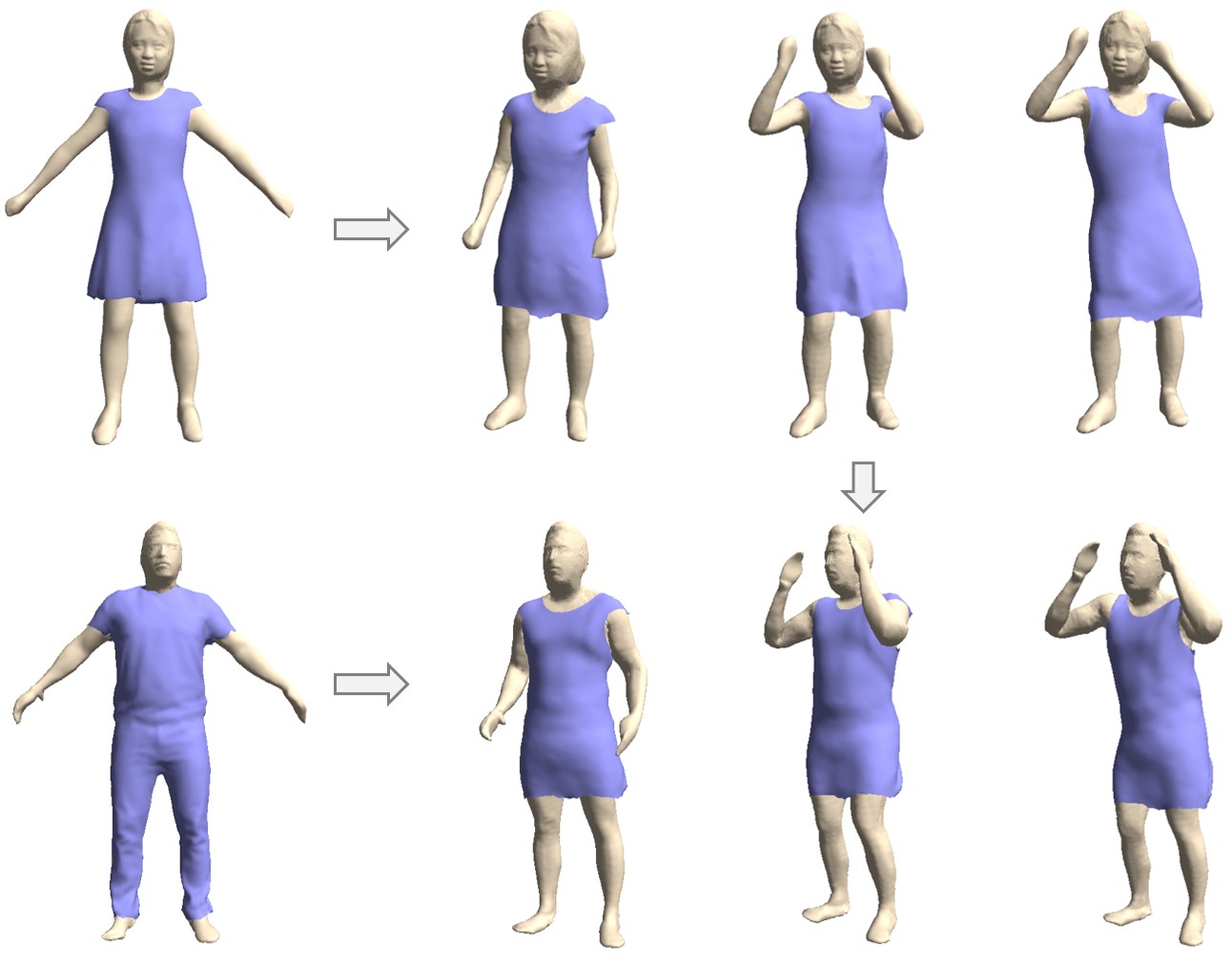}
  \vspace{1mm}
  \begin{picture}(0,0)
    \put(-225,-8){Reconstruction}
    \put(-110,-8){Animation results}
    \put(-60,82){clothing}
  \end{picture}
  \caption{Decoupled reconstruction applications.
  Reconstructed clothing and body can be animated using physical simulation methods.
  Additionally, clothing can be easily swapped to create animations with different outfits.
  }
  \label{change}
\end{figure}

\subsection{Quantitative Evaluation.}
Since there is no public real dataset available for evaluating the geometric quality of decoupled clothed human reconstruction from monocular video, we use four synthetic datasets provided by SelfRecon~\cite{jiang2022selfrecon}, each containing geometric ground truth and rendered videos.
As these processed data are not open source from REC-MV~\cite{qiu2023RECMV}, we manually labeled the split points and used the official tools to generate the feature lines.
We also employ CLO3D~\cite{clo3d} to separate clothing and body from the source data for quantitative evaluation.
We report the Chamfer Distance (CD) for clothing, body, and the complete clothed human of each method's results. 
For methods that do not support decoupling, we only report the CD for the complete clothed human.
We present the visualization of the quantitative comparison in Figure~\ref{compare_num} and report the metrics in Table~\ref{tab1}.
The results show that our method achieves the best results in the metrics, provides detailed and accurate visual effects, and is capable of correctly decoupling the human body and clothing.

\subsection{Qualitative Evaluation.}
We compare our method to methods capable of reconstructing clothed human bodies from image sequences, using several sequences from the PeopleSnapshot~\cite{alldieck2018videoavatar} dataset and one sequence from SelfRecon~\cite{jiang2022selfrecon} to include skirted clothing categories. 
For all methods, we extract consecutive frames from a complete rotation and present the comparison results for the first frame in Figure~\ref{compare2}.
The additional sequential results and discussion are presented in the supplementary materials.

As we can see, REC-MV~\cite{qiu2023RECMV} can accurately reconstruct clothing but lacks the level of detail achieved by our method. 
Due to the lack of further optimization for the body in REC-MV, directly using SMPL results in mesh penetration.
BCNet~\cite{jiang2020bcnet} reconstructs clothing that can directly use SMPL as the underlying body; however, it only supports reconstruction with consistent clothing categories and lacks almost all detail.
DELTA~\cite{Feng2023DELTA} enhances head details based on SCARF~\cite{Feng2022scarf}, and allows direct decoupled reconstruction of clothing and body.  
However, because it uses a NeRF~\cite{mildenhall2020nerf} representation for clothing, it cannot extract smooth geometry, leading to numerous artifacts in the clothing geometry.
SelfRecon~\cite{jiang2022selfrecon} uses the SDF representation to reconstruct clothed human bodies with correct shapes but lacking detail. 
GoMAvatar~\cite{wen2024gomavatar} employs a Gaussians-on-Mesh representation, resulting in relatively coarse meshes. 
Neither method can achieve decoupling of clothing from the body.
Compared to these methods, our approach successfully decouples clothing from the body while maintaining a richer level of detail.

\subsection{Ablation Study.}
\textbf{Implicit UDF or hmSDF?}
\label{sec:udf}
Several articles~\cite{liu2024gshell, de2023drapenet, Chen2024NeuralABC, long2022neuraludf} have demonstrated results by using the implicit Unsigned Distance Field (UDF) to represent clothing with undefined categories, leveraging dense supervision from meshes or multi-view images.
However, we found that UDF struggles to produce robust results due to the limited supervision provided by single-view dynamic human reconstruction, as shown in Figure~\ref{ablation_udf}.
The UDF reconstruction resulted in numerous small holes, a large hole in the abdominal area, and failed to create an opening at the cuffs. In contrast, hmSDF achieved an accurate garment shape.

Although UDF uses a network to model shapes and possesses strong representational capabilities, it still has certain problems: 
(1) UDF is non-differentiable at the 0-level set. Although several solutions~\cite{long2022neuraludf, guillard2022udf, Liu23NeUDF} have been proposed for mesh extraction and multi-view reconstruction problems, UDF remains sensitive near the 0-level set, and areas with poor supervision signals may lead to reconstruction failures.
(2) The ability to extract surfaces from implicit UDF is limited. Surface extraction from implicit UDF~\cite{guillard2022udf} is limited to manifold surfaces~\cite{lorensen1987marching}, and regions that are non-manifold in the implicit representation may fail to be extracted.
(3) The strong representational capability of UDF reduces its noise resistance; for instance, the large hole in the abdominal area is caused by occlusion in the clothing mask. An example of occlusion is shown in Figure~\ref{mask_occ}.

\textbf{Perceptual Normal Loss.}
We attempt to remove the normal consistency loss and replace it with mean squared error and angular error. 
While the normal consistency loss focuses on feature consistency and can produce perceptually consistent results, 
the mean squared error and angular error focus on pointwise features, which may result in less smooth results. 
We show the normal rendering results in Figure~\ref{ablation_normal}.
Calculating the cosine or MSE loss between the rendered normals and the reference normals results in rough and noisy reconstructions. 
Using the perceptual normal loss, on the other hand, produces smoother results that preserve the features and details of the reference image.

\subsection{Applications.}
We demonstrate the clothing transfer after decoupled reconstruction and the application of physics-based animation.
The results are shown in Figure~\ref{change}.

\textbf{Clothing Transfer.}
Since our model is capable of reconstructing decoupled clothing and human bodies, clothing transfer can be achieved by separately reconstructing two clothed human figures and exchanging their garments. 

\textbf{Physics-based Animation.}
The reconstructed clothing and body geometry can be used with physics-based simulation methods to create more realistic animations. 
Compared to non-decoupled reconstructions~\cite{jiang2022selfrecon}, which make it difficult to establish motion between clothing and body, we use HOOD~\cite{grigorev2022hood} to create more authentic clothing details.

\section{Conclusion}

We introduced $\text{D}^3\text{-Human}$, a method that could reconstruct decoupled clothing and body directly from a short monocular video. By leveraging the robustness of explicit representations and the flexibility of implicit representations, $\text{D}^3\text{-Human}$ ensured the reconstruction of detailed features while maintaining the plausibility of body parts obscured by clothing. To achieve the separation of 3D clothing from the body, we proposed a novel representation called hmSDF defined on human body, which was able to obtain 3D segmentation using only 2D human parsing, without any 3D clothing priors.
Thanks to this novel approach, we are able to achieve competitive reconstruction accuracy with much less computation time while ensuring the decoupling of clothing and body. 
The decoupled reconstruction results can be easily used for detailed animation production and clothing transfer. 
Our $\text{D}^3\text{-Human}$ can create high-quality and easily editable human geometry using only one camera, providing a technical foundation for the widespread adoption of many applications, such as highly editable digital avatar creation, holographic communication.

% \input{sec/X_suppl}
% \clearpage
{
    \small
    \bibliographystyle{ieeenat_fullname}
    \bibliography{main}
}

\end{document}